\documentclass[letterpaper, 10 pt, conference]{ieeeconf}  

\IEEEoverridecommandlockouts                              

\usepackage{amsmath} 
\usepackage{array}
\usepackage{booktabs} 
\usepackage{amsmath}
\usepackage{amssymb}
\usepackage{tabularx}
\usepackage{graphicx}
\usepackage{multirow}
\usepackage{graphicx}

\title{\LARGE \bf
HeLoM: Hierarchical Learning for Whole-Body \\ Loco-Manipulation by a Hexapod Robot
}

\author{
Xinrong Yang$^{1*}$, Peizhuo Li$^{1*}$, Hongyi Li$^{1,2}$, Yifeng Peng$^{1}$, Arhaan Jain$^{1}$, Junkai Lu$^{1}$, Linnan Chang$^{1}$,\\ Yuhong Cao$^{1}$, Yifeng Zhang$^{1\dagger}$, Ge Sun$^{1}$, Guillaume Sartoretti$^{1}$ 
\thanks{$^{*}$ These authors contributed equally to this work.}
\thanks{$^{\dagger}$ Corresponding author: Yifeng Zhang (yifengz@nus.edu.sg)}
\thanks{$^{1}$ MARMoT Lab, Department of Mechanical Engineering, National University of Singapore, Singapore}
\thanks{$^{2}$ Center for X-mechanics, Zhejiang University, Hangzhou, China}
}

\begin{document}

\maketitle
\thispagestyle{empty}
\pagestyle{empty}

\begin{abstract}
In nature, animals often need to move/manipulate objects comparable in weight/size to their own bodies.
Compared to grasping and carrying, pushing provides a more straightforward and efficient non-prehensile manipulation strategy, avoiding complex grasp design while leveraging direct contact to regulate an object's pose during interaction.
Achieving effective pushing, however, requires both sufficient manipulation capability and stable whole-body coordination, which is particularly challenging when dealing with heavy or irregular objects. 
To address these challenges, we propose \textbf{HeLoM}, a learning-based hierarchical whole-body manipulation framework for hexapod robots that exploits coordinated multi-limb control and is applicable to multi-legged robotic systems.
Inspired by the cooperative strategies of multi-legged insects, our framework leverages multiple contact points and high degrees of freedom to enable efficient and dynamic whole-body coordination during object interaction. 
HeLoM's high-level planner plans pushing behaviors, while its low-level controller maintains locomotion stability and generates dynamically consistent joint actions.
This design enables the robot to maintain balance while executing continuous and controllable pushing behaviors through coordinated foreleg interaction and supportive hind-leg propulsion. 
We validate the effectiveness of \textbf{HeLoM} through both simulation and real-world experiments.
Results show that our framework can stably push objects of varying sizes and unknown physical properties to designated goal poses in the real world.

\end{abstract}



\section{INTRODUCTION}

Mobile manipulation addresses the integration of a dedicated mobile base with manipulation capabilities, allowing robots to perform tasks that demand coordinated movement as well as dexterous interaction~\cite{gong2023legged}.
Transporting heavy objects, such as delivery packages or toolboxes comparable in weight to the robot demands flexible and well-coordinated mobility and manipulation capabilities.
In such scenarios, non-prehensile manipulation (e.g., pushing or sliding) often provides a more straightforward and capable solution than prehensile manipulation (e.g., grasping or carrying)~\cite{dengler2022learning}, for many tasks that allow it.
This approach eliminates the need for complex gripper design to instead rely on direct contact between the robot's body or limbs and the object to regulate its pose during movement.
Legged robots are particularly well-suited for this form of manipulation.
With their high degrees of freedom, they can coordinate limb motions through whole-body control to achieve efficient and continuous pushing behaviors throughout the task~\cite{jeon2023learning}.

With the rapid development of learning-based locomotion and posture control in legged robots~\cite{peizhuo2025sata}, researchers have begun extending these methods to manipulation tasks in order to achieve whole-body manipulation~\cite{portela2025whole}.
Early approaches relied on using the robot's body to directly contact and apply force to an object~\cite{jeon2023learning}.
While this strategy is simple to implement, it often requires frequent adjustments of the robot's body position, resulting in inefficient transport and prolonged task duration. 
Moreover, impact forces are directly transmitted to the robot's actuators and sensors, posing risks of damage and reducing long-term reliability.
To mitigate these issues, some studies introduced dedicated robotic arms, which enable more precise interactions and flexible force application through end effectors~\cite{dadiotis2025dynamic}.
While this approach improves manipulation dexterity, contact is generally restricted to a single point, limiting the ability to deliver continuous and stable pushing forces to large objects.
In addition, the inclusion of an arm inevitably increases system complexity, design cost, and control difficulty.
Researchers have also explored pushing through direct contact with the feet (end effectors) of quadruped robots~\cite{cheng2025rambo}.
This method requires no additional hardware and naturally integrates locomotion with manipulation.
However, with fewer feet grounded during pushing, the robot's overall stability is reduced, and the manipulation capability of individual legs becomes limited, which constrains the effectiveness of this approach.

\begin{figure}
    \vspace{0.2cm}
    \centering
    \includegraphics[width=\linewidth]{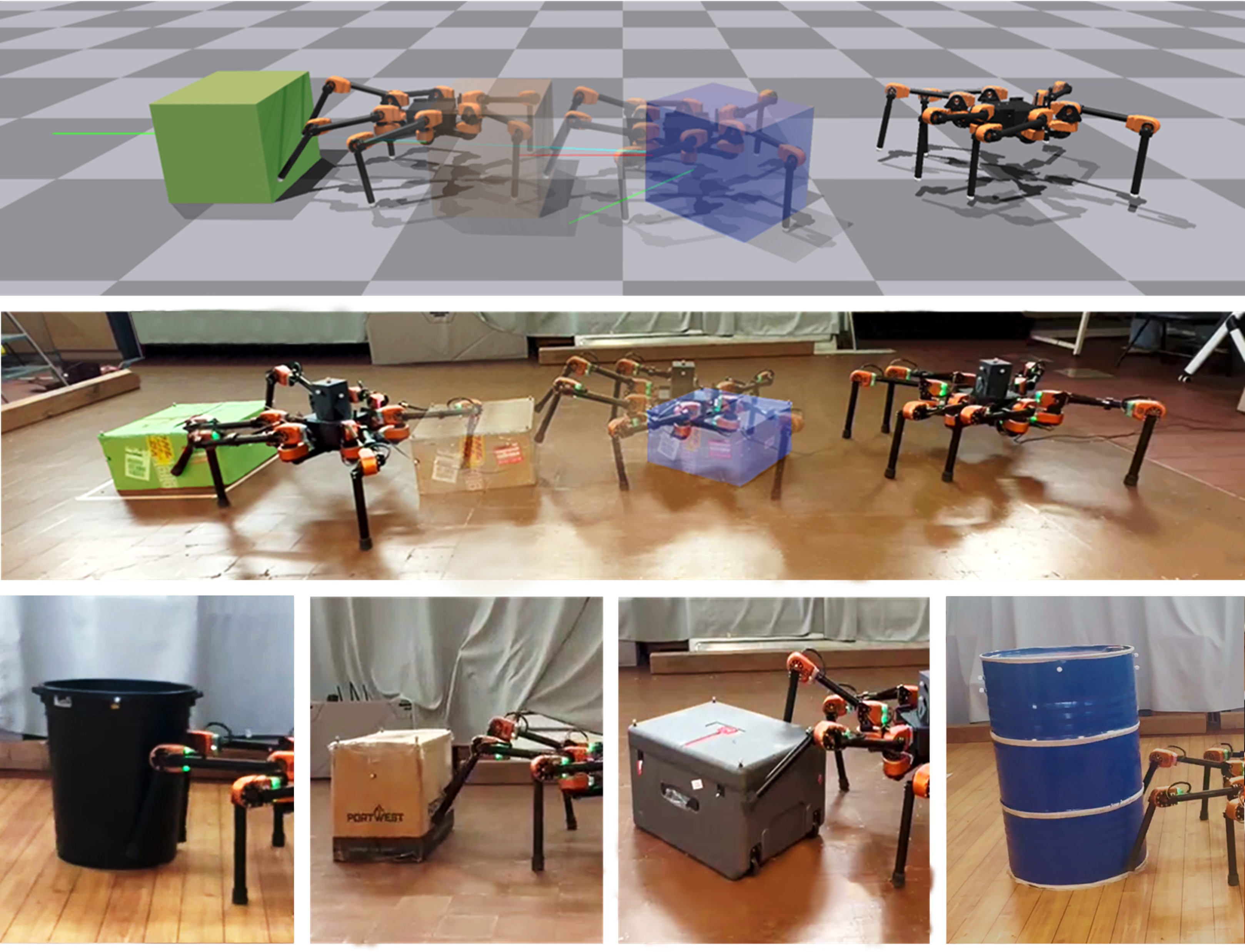}
    \vspace{-0.75cm}
    \caption{Examples of our series-elastic hexapod robot pushing different objects to desired poses using its forelegs with our proposed \textbf{HeLoM} framework, both in simulation (top) and real-world experiments (bottom).}
    \vspace{-0.6cm}
    \label{fig:front_page}
\end{figure}

Biological inspiration suggests that robots with more than four legs may offer clear advantages for manipulation.
Insects such as hexapods and octopods exhibit remarkable coordination strategies when interacting with objects~\cite{leung2024nature}. 
They can use a subset of their legs to maintain simultaneous contact, continuously adjust object orientation, and move at the same time, while the remaining legs provide balance and propulsion.
In tasks such as dung-ball rolling, beetles further coordinate their whole body to maximize force application. 
Such multi-limb cooperation leverages redundant degrees of freedom and contact points to improve stability, precision, and manipulation performance.

Inspired by the manipulation strategies observed in multi-legged insects, we propose \textbf{HeLoM}, a reinforcement learning (RL)-based control framework for hexapod robots that can be extended to other multi-legged robotic systems, which exploits whole-body coordination to accomplish large object pushing tasks.
Our framework adopts a hierarchical structure consisting of a \textit{Planner} and a \textit{Controller}.
Our \textit{Planner} directs commands to the \textit{Controller} to drive coordinated dual-foreleg pushing behaviors, with the goal of moving the object progressively closer to its target position and orientation.
Meanwhile, our \textit{Controller} focuses on accurately executing these commands by coordinating all joints to maintain balance, handle external disturbances, and optimize torque distribution, as illustrated in Fig~\ref{fig:front_page}.
Through this process, the robot achieves stable whole-body movements and effectively executes pushing behaviors, ensuring that the \textit{Planner}'s intended manipulation strategy is faithfully realized.
In doing so, our work demonstrates the potential of hexapod robots in challenging manipulation tasks and offers a promising pathway toward whole-body manipulation with multi-legged platforms.
Thus, our main contributions are:

\begin{itemize}
    \item We propose \textbf{HeLoM}, a bio-inspired hierarchical RL framework for whole-body loco-manipulation in multi-legged systems. Our approach reformulates loco-manipulation as learning whole-body interaction behaviors, instead of decoupling locomotion and manipulation. To the best of our knowledge, it is the first learning-based realization of whole-body loco-manipulation on a large hexapod robot, supporting more complex manipulation tasks in multi-legged robots.

    \item We train a high-level \textit{Planner} and a low-level \textit{Controller}, where the \textit{Planner} outputs commands to the \textit{Controller}, which unifies locomotion and manipulation to realize coordinated whole-body behaviors. By leveraging the high degrees of freedom and multiple contacts of multi-legged systems, our approach enables efficient and continuous pushing without requiring prior knowledge of object properties.

    \item We validate the effectiveness and robustness of our approach through extensive experiments on a real hexapod robot, addressing the sim-to-real gap and demonstrating policy generalization to objects of varying sizes, shapes,and out-of-distribution (OOD) scenarios.
\end{itemize}


\section{Related Works}

\subsection{Legged Manipulation Controller}

Recent interest in legged robots has expanded from locomotion to manipulation, leading to the exploration of various control approaches. Optimal control has been applied to legged robot pushing, including pushing with the body~\cite{sombolestan2022hierarchical}, pushing with a single dedicated arm~\cite{xin2022loco, ferrolho2023roloma}, and pushing with two arms of a centaur-like robot~\cite{polverini2020multi}. However, these optimal control-based approaches rely on precise object models, which must be assumed or estimated through costly and often inaccurate identification. In addition, convexifying complex loco-manipulation problems typically requires simplifying robot dynamics~\cite{di2018dynamic}, introducing control errors and limiting versatility in contact-rich tasks.
To address these limitations, recent work has turned to fully RL-based controllers for legged manipulation. Ji et al. trained a dynamic loco-manipulation policy that enables a quadruped to dribble a ball while running over rough terrain~\cite{ji2023dribblebot}. Other studies focus on legged robots with dedicated arms: Fu et al. and Portela et al. developed whole-body controllers on arm-mounted quadruped platforms, achieving manipulation in the position and force domains, respectively \cite{fu2023deep, portela2024learning}. Pushing the boundaries of limb utilization, Zhu et al. introduced the ReLIC framework to achieve flexible interlimb coordination, enabling the quadruped to dynamically assign its legs for either locomotion or assisting the arm in manipulation \cite{zhu2025relic}. Building upon such complex loco-manipulation systems, Ha et al. further enhanced their autonomy by introducing a task-centric high-level policy \cite{ha2024umi}.

Nevertheless, most prior work focuses on quadrupeds, while multi-legged robots remain underexplored. With more legs, hexapod robots provide enhanced stability under external reaction forces in contact-rich tasks~\cite{sun2022joint}, even when two legs are engaged in manipulation. Existing studies, such as using B\'{e}zier trajectories for obstacle manipulation~\cite{lu2020autonomous} and hexapods grasping objects~\cite{xu2024design}, generally treat the body as a fixed base and overlook body–leg coordination, leaving much of their potential for agile loco-manipulation untapped. 
While a recent multitask framework achieves simultaneous loco-manipulation~\cite{zhang2024legs}, it still lacks autonomous planning capabilities.

\subsection{Legged Robot Pushing}
\begin{figure*}[t]
    \centering
    \includegraphics[width=1.0\textwidth]{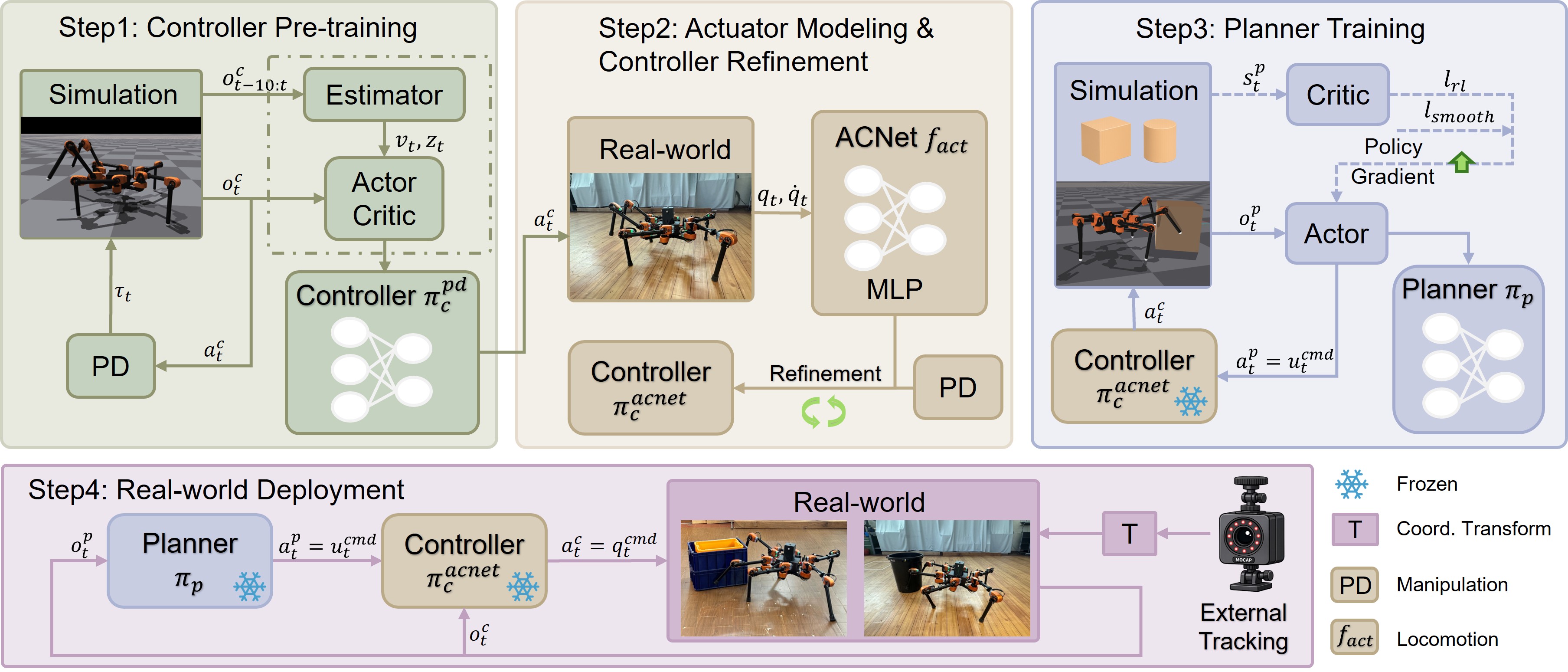}
    \vspace{-0.6cm}
    \caption{\textbf{Overview of the proposed HeLoM framework}. During training, (1) we first pre-train the \textit{Controller} with randomly sampled commands and a PD controller to establish basic loco-manipulation capabilities. (2) We then collect real-world data using the pre-trained controller, train an actuator network, and replace the PD controller to further refine the \textit{Controller} with learned actuator dynamics. (3) Next, the \textit{Controller} is frozen, and the \textit{Planner} is trained on top of it to generate effective pushing commands. To improve smoothness and coordination, our \textit{Planner} and \textit{Controller} are executed synchronously at 50 Hz. (4) During deployment, a motion-capture system provides the robot and object states in the world frame, followed by coordinate transformation.
}
    \label{fig:system}
    \vspace{-0.4cm}
\end{figure*}

Research on legged robot pushing has investigated a variety of approaches. Rigo et al. employed hierarchical MPC to optimize contact forces and locomotion for planar box pushing with the robot body, but their method relies on accurate object models and flat terrain~\cite{rigo2022contact}. Improving on this, Sombolestan and Nguyen incorporated online estimation of unknown mass and friction, enabling quadrupeds to push objects on slopes; however, their framework only models the robot's dynamics and does not explicitly regulate the object's pose~\cite{sombolestan2022hierarchical}. Jeon et al. advanced body-based pushing by training whole-body repositioning policies through hierarchical control and deep latent embeddings, achieving high success rates even with heavy loads. Yet, their controller's effectiveness remains constrained by the body-pushing paradigm, which limits the robot's workspace and the directions in which forces can be applied~\cite{jeon2023learning}. Recognizing the drawbacks of body pushing, Arm et al. shifted the focus to leg-based pushing, where a robust RL controller enables individual legs to perform diverse non-prehensile tasks. This approach expands the contact point's workspace, nevertheless, high-level trajectory planning still depends on human teleoperation~\cite{arm2024pedipulate}. 
Dadiotis et al. proposed a constrained RL policy as a high-level controller for quadrupeds equipped with arms, allowing adaptive selection of contact points to prevent toppling~\cite{dadiotis2025dynamic}. However, their method requires multiple pushes to achieve object repositioning and reorientation. 
To overcome the reliance on repeated contacts, Hong et al. introduced a multi-agent system in which two quadrupeds collaborate to push large objects~\cite{feng2025learning}. 
Although effective, the cost of multiple robots and the reliability of robot communication remain a concern. In a different direction, Leung et al. drew inspiration from dung beetles, adapting hexapod leg movements to roll balls across varied terrains. Nonetheless, their study focuses on ball rolling and does not generalize to broader planar pushing tasks~\cite{leung2024nature}.


\section{Hierarchical Loco-Manipulation Learning}

Our framework addresses the pushing task for hexapod robots by decomposing the problem into interaction planning and control execution. 
Both layers are formulated as RL problems:
The \textit{Planner} learns to produce pushing commands, while the \textit{Controller} learns to translate them into feasible whole-body joint actions under contact constraints. An overview of the \textbf{HeLoM} architecture is provided in Fig.~\ref{fig:system}.


\subsection{Pushing Planner}

Our \textit{Planner} determines the desired trajectories of the two front feet (end-effectors) along with  body motions to generate effective interaction behaviors, guiding the object to its target position and orientation (yaw). Its RL formulation is introduced as follows:

\subsubsection{Observations \& Action Space}

\begin{table}[t]
\centering
\caption{Planner Observation Representation and Notations.}
\vspace{-0.3cm}
\label{tab:obs_notation}
\renewcommand{\arraystretch}{1.3}
\resizebox{\columnwidth}{!}{%
\begin{tabular}{|c|c|c|c|c|c|c|}
\hline
\multicolumn{2}{|c|}{\textbf{Notations}} & \textbf{Components} & \textbf{Dim} & \textbf{Components} & \textbf{Dim} & \textbf{Total} \\ \hline

\multirow{8}{*}{$s_{t}^{p}$} 
& \multirow{3}{*}{$o_{t}^{p}$} 
& $ {p}_{object} $ & 3  & $ {p}_{object}^{cmd} $  & 3  & \multirow{3}{*}{81}  \\ \cline{3-6}

&  & $ {R}_{object} $ & 3  & $ {R}_{object}^{cmd} $ & 3  & \\
\cline{3-6}
& &  $ o_{t}^{c} $ & 69 & $ $  &  & \\
\cline{2-7}

& \multirow{5}{*}{$x_{t}^{p}$}
 & $dimensions$ & 3 & $e_{object}$ & 1 & \multirow{5}{*}{20} \\ \cline{3-6}
& & $m_ {object}$ & 1 & $contact$ & 2 & \\ \cline{3-6}
& & $  {com} $ & 3 & $\mu_ {object}$ & 1  & \\ \cline{3-6}
& & $ {I}_ {object}$ & 9 &  &  &   \\ \hline
\end{tabular}
}
\vspace{-0.2cm}
\end{table}

\begin{table}[t]
\centering
\caption{Planner reward terms and weights ($dt=0.02$).}
\vspace{-0.3cm}
\label{tab:planner_reward}
\renewcommand{\arraystretch}{1.2}
\begin{tabular}{>{\centering\arraybackslash}p{0.2\linewidth}
                >{\centering\arraybackslash}p{0.50\linewidth}
                >{\centering\arraybackslash}p{0.15\linewidth}}
\toprule
\textbf{Term} & \textbf{Expression} & \textbf{Weight} \\
\midrule
 
$r_{\text{foot,obj}}$
& $\frac{1}{2}\sum_{i \in \{l,r\}} 
\exp\left(-{d_i}/{\sigma_{fo}}\right)$ 
& $1dt$ \\

$r_{\text{foot,contact}}$ 
& $\frac{1}{2}\sum_{i \in \{l,r\}} 
\exp\left(-{[h_i - h^*]_+}/{\sigma_{fh}}\right)$ 
& $0.8dt$ \\

$r_{\text{obj,tar}}$ 
& $\alpha\exp\!\left(-\|p_{obj}-p_{tar}\|/\sigma_{tn}\right)
+ (1-\alpha) \left(1 + \|p_{obj}-p_{tar}\|/(d_{task}\sigma_{tf})\right)^{-1}$ 
& $3dt$ \\

$r_{\text{obj,dir}}$ 
& $(v_{obj}^{xy} \cdot \hat{d}_{goal})\, \lambda_{\text{push}}$ 
& $1.5 dt$ \\

$r_{\text{obj,yaw}}$
& $\exp\!\left(-\theta_z^2/\sigma_{\theta}^2\right)
\, \lambda_{\text{turn}}$
& $1 dt$ \\

$r_{\text{obj,turn}}$
& $\mathrm{ReLU}(\omega_z \tanh(e_{\theta}^{s}/\sigma_{e}))\, \lambda_{\text{turn}}$
& $0.4 dt$ \\

$r_{\text{obj,tilt}}$ 
& $\|g_{obj}^{xy}\|^2$ 
& $-10dt$ \\

$r_{\text{termination}}$ & $\mathbb{I}_{\mathrm{finish}} $ & $10dt$ \\

$r_{\text{action rate}}$ & $ \|a_t^{p}-a_{t-1}^{p}\|^2 $ & $-0.5dt$ \\

$r_{\text{action limits}}$ & $ \sum_j \left([a_j^{\min}-a_{t,j}]_+ + [a_{t,j}-a_j^{\max}]_+\right)$ & $-1dt$ \\

\bottomrule
\end{tabular}
\vspace{-0.4cm}
\end{table}

To endow our \textit{Planner} policy with planning capability that accounts for the robot's posture, we design the observation vector $o_{t}^{p}$. As summarized in Table~\ref{tab:obs_notation}, this input not only includes the object pose in the base frame $p_{object} \in \mathbb{R}^3,\, R_{object}\in \mathbb{R}^3$, and the target pose $p_{object}^{cmd} \in \mathbb{R}^3,\, R_{object}^{cmd} \in \mathbb{R}^3$, but also incorporates the full observation vector $o_{t}^{c}$ used by the low-level \emph{Controller}, where $o_{t}^{c}$ is defined as: 
\begin{equation}
o_{t}^{c} = [\omega_t,\, g_t,\, q_t,\, \dot{q}_t,\, a_{t-1}^{c},\, u_{t}^{cmd}]^T.
\label{eq:controller_obs}
\end{equation}
Here, $\omega_t \in \mathbb{R}^3$ denotes the base angular velocity, and $g_t \in \mathbb{R}^3$ is the gravity vector. The vectors $q_t \in \mathbb{R}^{18}$ and $\dot{q}t \in \mathbb{R}^{18}$ represent the joint positions and velocities measured by the encoders, while $a_{t-1}^{c} \in \mathbb{R}^{18}$ corresponds to the previous action of the \textit{Controller}. The command vector $u_{t}^{cmd} \in \mathbb{R}^9$, which also serves as the previous output (i.e., action) $a_{t-1}^{p}$ of the \textit{Planner} actor network, includes both the desired base velocity $v_t^{\text{cmd}} = [v^{cmd}{x,t}, v^{cmd}{y,t}, \omega^{cmd}_{yaw,t}]$ and the desired positions of the two front feet in the base frame $p_t^{\text{cmd}} \in \mathbb{R}^6$.

In real-world settings, the physical properties of the manipulated object, such as its mass $m_{object}$, $dimensions$, center of mass $com$, inertia matrix ${I_{objec}}$, coefficient of restitution $e_{objec}$, and surface friction coefficient $\mu_{objec}$, are typically unknown. 
Therefore, we treat these properties as privileged observations and provide them only to the critic network during training. Furthermore, we find that incorporating a binary indicator $contact$ into the critic network, which represents whether the two front feet are in contact with the object, improves the accuracy of value estimation. We empirically observed that this improvement provides more reliable feedback to the actor and facilitates more effective exploration during the early stages of training.

\subsubsection{Reward Functions}

To enable the robot to push various objects to target poses using its feet, we design a reward function composed of four groups of terms, which we describe below and in Table~\ref{tab:planner_reward}: \\[-0.3cm]

\noindent \textbf{Foot-object interaction:} 
$r_{\text{foot,obj}}$ drives the feet toward the nearest surface of the object based on a distance metric. Compared to approaches that target the object center, this design allows the feet to adaptively adjust contact locations during interaction, enabling more precise and stable placement. 
$r_{\text{foot,contact}}$ encourages contacts at lower regions of the object, which improves pose stability and reduces the risk of tipping, especially under high-friction or low center-of-mass conditions. Furthermore, to enhance training stability, the environment is reset when the object's tilt angle around the $x$ and $y$ axes exceeds $40^\circ$, thereby filtering out unstable states and accelerating convergence. \\[-0.3cm]

\noindent \textbf{Goal-directed object motion:} 
$r_{\text{obj,tar}}$  guides the object toward the target position. Considering that the target distance may vary over a wide range, we design two complementary distance-based terms to provide effective gradients at different stages. Specifically, $\exp\!\left(-\|p_{obj}-p_{tar}\|/\sigma_{tn}\right)$ provides strong gradients when the object is close to the target, facilitating precise convergence, while $\left(1 + \|p_{obj}-p_{tar}\|/(d_{task}\sigma_{tf})\right)^{-1}$ provides a smoother and persistent signal at larger distances, reflecting the task progress and reducing inefficient exploration in early stages. $r_{\text{obj,dir}}$ encourages the object to move along the direction toward the target, guiding the robot to approach the goal along near-optimal paths and improving overall motion efficiency. In addition, $r_{\text{obj,yaw}}$ and $r_{\text{obj,turn}}$ jointly regulate the object orientation, ensuring that the desired pose is achieved upon reaching the target. In particular, $r_{\text{obj,turn}}$ provides stronger rotational guidance under large angular deviations, compensating for the diminished gradients of $r_{\text{obj,yaw}}$ in such cases, thereby mitigating unstable exploration when the orientation error is large.

Furthermore, we introduce weighting coefficients $\lambda_{\text{turn}}$ and $\lambda_{\text{push}}$, defined as:
\begin{equation}
\lambda_{\text{turn}} = \mathrm{clip}(\theta_z/\sigma_1,\, 0,\, 1),\quad
\lambda_{\text{push}} = \exp(-(\theta_z/\sigma_2)^2)
\end{equation}
where $\sigma_1 = 0.5$ and $\sigma_2 = 0.8$. These coefficients are designed as smoothly varying functions to enable a gradual transition from rotation-dominant to translation-dominant behaviors during training. 
If rotation and translation are optimized simultaneously with equal emphasis, the rotation phase may temporarily increase the distance to the target, thereby weakening distance-based rewards and leading to suboptimal local solutions. \\[-0.3cm]

\noindent \textbf{Regularizing manipulation behavior:} 
$r_{\text{orientation}}$ penalizes undesired object rotations around the $x$ and $y$ axes, which are often caused by unstable contacts or collisions. By constraining the object's orientation, this term implicitly encourages smoother and more controlled pushing behaviors, reducing impulsive interactions and unnecessary contacts. Furthermore, it improves sim-to-real transfer by mitigating unstable contact patterns, leading to more predictable and robust manipulation in real-world scenarios. $r_{\text{action rate}}$ and $r_{\text{action limits}}$ encourage smooth actions and regular control behavior, reducing abrupt changes in joint commands and helping the robot maintain stable and predictable motions throughout the pushing process. \\[-0.3cm]

\noindent \textbf{Terminal reward:} When the object's positional error is less than 0.05\,m and its orientation error within $5^\circ$, the task is considered a success. A fixed terminal reward $r_{\text{termination}}$ of $+10.0$ is then applied to provide a stronger learning signal, amplifying the policy gradient for successful steps. 


\subsection{Whole-Body Loco-manipulation Controller}

We adopt the DreamWaQ~\cite{nahrendra2023dreamwaq} architecture for our \textit{Controller}, enabling smooth tracking of the \textit{Planner}'s commands. In particular, we leverage its estimator module to encode a history of proprioceptive observations into a compact latent representation. We detail our RL formulation below:

\subsubsection{Observations \& Action Space}

The \textit{Controller} actor network takes as input the observation vector $o_{t}^{c}\in \mathbb{R}^{69}$, together with latent variables $v_t\in \mathbb{R}^{3}$ and $z_t\in \mathbb{R}^{16}$, which are outputs of the estimator given a proprioceptive history $o_{t-10:t}^{c}$. The estimator is trained jointly with the policy in an end-to-end manner. There, $v_t$ represents the estimated base linear velocity, while $z_t$ represents a latent variable for understanding terrain properties. The actor finally outputs the desired joint positions $a_{t}^{c} \in \mathbb{R}^{18}$.

\subsubsection{Reward Functions} 

\begin{table}[]
\centering
\caption{Controller reward terms and weights.}
\vspace{-0.3cm}
\label{tab:controller_reward}
\renewcommand{\arraystretch}{1.2}
\begin{tabular}{>{\centering\arraybackslash}p{0.18\linewidth}
                >{\centering\arraybackslash}p{0.46\linewidth}
                >{\centering\arraybackslash}p{0.22\linewidth}}
\toprule
\textbf{Reward} & \textbf{Expression} & \textbf{Weight} \\
\midrule
 $r_{\text{tracking},xy}$ & $\exp\!\left(-\| v^{{cmd}}_{xy} - v_{xy} \|^2 / \sigma_{xy} \right)$ & $2dt$ \\
$r_{\text{tracking},z}$ & $\exp\!\left(-\| \omega^{{cmd}}_{z} - \omega_z \|^2 / \sigma_{z} \right)$ & $1dt$ \\
$r_{\text{tracking},p}$  & $\sum_{i \in \{l,r\}} \exp\!\left(-\| p^{cmd}_i - p_{i}\|^2 / \sigma_{p}\right)$ & $2dt$ \\

$r_\text{leg\ lift}$       & $\sum_{i \in \{l,r\}}\mathbb{I}(F_{z,i} < F_{\text{lift}})$ & $0.5dt$ \\
$r_\text{joint\ deviation}$ & $ \| q_t - q_{default} \|^2 $& $-0.3dt$ \\
$r_\text{joint\ acc}$& $\left\| (\dot{q}_t - \dot{q}_{t-1})/dt\right\|^2$ & $-2.5\times 10^{-7}dt$ \\
$r_\text{torque}$  & $\tau_t^2$ & $-1\times 10^{-5}dt$ \\
$r_\text{action\ rate}$  & $ \|a_t^{c}-a_{t-1}^{c}\|^2 $ & $-0.03dt$ \\
$r_\text{velocity\ penalty}$  & $v_z^2$ & $-2dt$ \\
$r_\text{orientation}$  & $\| g_{xy} \|^2$ & $-20dt$ \\
\bottomrule

\end{tabular}
\vspace{-0.6cm}
\end{table}

To ensure stable execution of our \textit{Controller}, we design the reward structure summarized in Table~\ref{tab:controller_reward}. The rewards terms are grouped into two categories: \\[-0.3cm]

\noindent \textbf{Tracking:} 
$r_{\text{tracking},xy}$ and $r_{\text{tracking},z}$ encourage the robot to follow the desired velocity, while $r_{\text{tracking},p}$ and $r_\text{leg\ lift}$ guide the two front feet toward their respective target positions. \\[-0.3cm]

\noindent \textbf{Regularization:} 
$r_\text{joint\ deviation}$ promotes the rapid development of well-structured motion patterns during early training, which accelerates policy convergence. The remaining reward terms collectively encourage stable body posture, smoother joint motions, and energy-efficient behavior. These regularization terms improve robustness and control stability during both locomotion and manipulation.


\subsection{Training Details}
We train our \textit{Controller} and \textit{Planner} using the proximal policy optimization (PPO)~\cite{schulman2017proximal}, a widely used RL algorithm that supports stable and efficient policy optimization.

\subsubsection{Controller} 

To avoid unstable behaviors caused by ill-posed target commands during training, we impose explicit constraints on both the velocity commands and the target foot positions. 
The commanded body velocity is bounded within $[-0.5,\,0.5]$~m/s along the $x$ axis, $[-0.3,\,0.3]$~m/s along the $y$ axis, and $[-1,\,1]$~rad/s in yaw. 
The target positions of the two front feet are constrained around their default poses. For each foot, the offset from its default pose is limited to $[-0.25, 0.25]$ m, $[-0.2, 0.2]$ m, and $[-0.2, 0.2]$ m along the x-, y-, and z-axes of the body frame, respectively.
Additionally, to prevent ill-posed or overlapping references, we enforce the constraint
$p_{\mathrm{rf},y} < p_{\mathrm{lf},y} - 0.15$,
ensuring sufficient separation between the right and left foreleg targets along the $y$ axis. We apply domain randomization over key physical parameters, as summarized in Table~\ref{tab:domain_rand}. The nominal gains for $k_p$ and $k_d$ are set to 100 and 2.0, respectively.

 It is worth noting that our physical robot is equipped with Series Elastic Actuators (SEAs), in which torque is transmitted through an elastic element placed in series with the drivetrain. This introduces compliance and additional actuator dynamics (e.g., spring-mass behavior and force control loops) that are not captured by simple PD-based rigid-actuator approximations, thereby increasing the sim-to-real gap. To improve robustness in the real system and mitigate this gap, we first train the policy using randomized settings in Table~\ref{tab:domain_rand}. The actuator data is then collected from comprehensive locomotion-only roll-outs to avoid ambiguity arising from contact-dependent torque variations. A total of 12 actuator-specific datasets are obtained for the 4 legs undergoing complex locomotion, and used to train 12 independent actuator networks. Consistent with the methodology of Hwangbo et al.~\cite{hwangbo2019acnet}, each actuator network is implemented as a supervised MLP that maps short histories of joint position error and velocity to the corresponding measured torque. These learned networks then replace the PD-based control for the locomotion legs during policy training, improving the fidelity of actuator dynamics in simulation and enabling more reliable sim-to-real transfer.

\begin{table}[t]
\centering
\caption{Domain randomization parameters.}
\vspace{-0.3cm}
\label{tab:domain_rand}
\renewcommand{\arraystretch}{1.1}
\begin{tabular}{c c c}
\toprule
\textbf{Parameter} & \textbf{Range} & \textbf{Unit} \\
\midrule
Friction coefficient & $U(0.4,\,1.2)$ & -- \\
COM shift $x$        & $U(-0.2,\,0.1)$ & m \\
COM shift $y$        & $U(-0.1,\,0.1)$ & m \\
COM shift $z$        & $U(-0.1,\,0.1)$ & m \\
Link mass scale      & $U(0.9,\,1.1)$ & -- \\
Inertia scale        & $U(0.9,\,1.1)$ & -- \\
$k_p, k_d$ scale     & $U(0.9,\,1.1)$ & -- \\
\bottomrule
\end{tabular}
\vspace{-0.2cm}
\end{table}


\subsubsection{Planner}
\label{sec:planner_training}

\begin{table}[]
\centering
\caption{Object parameter randomization ranges.}
\vspace{-0.3cm}
\label{tab:obj_size}
\renewcommand{\arraystretch}{1.1}
\setlength{\tabcolsep}{4pt}
\begin{tabular}{lcclcc}
\toprule
\textbf{Parameter} & \textbf{Range} & \textbf{Unit} &
\textbf{Parameter} & \textbf{Range} & \textbf{Unit} \\
\midrule
Length  & $U(0.3,\,0.5)$   & m &
Mass & $U(5.0,\,15.0)$ & kg\\
Width   & $U(0.25,\,0.5)$ & m &
Friction & $U(0.3,\,0.8)$ & -- \\
Height  & $U(0.45,\,0.60)$  & m &
Object type & $\text{[Box, Cylinder]}$ & -- \\
\bottomrule
\end{tabular}
\vspace{-0.5cm}
\end{table}

During training of the \textit{Planner}, we freeze the parameters of the \textit{Controller} to ensure reliable execution of motor commands. We randomize the initial poses of both the robot and the object, as well as the target object poses, within a $6\,\text{m} \times 6\,\text{m}$ workspace. Each parallel environment is equipped with a versatile object generator that randomizes physical properties, as summarized in Table~\ref{tab:obj_size}, with boxes and cylinders sampled at a 1:1 ratio. The ground friction coefficient is uniformly sampled from $[0.3,\,1.0]$.

To reduce the exploration difficulty in early training, we adopt a command-based curriculum that progressively increases task complexity. Directly sampling targets from the full workspace and orientation range often leads to ambiguous learning signals, causing the robot to remain stationary after contact or fail in long-range pushing and large-angle alignment. Specifically, we discretize the target displacement along the $y$-axis (within a $6\,\text{m}$ range) and the angle (within $\pm 180^\circ$) into 5 curriculum levels. The curriculum levels for translation ($y$) and orientation (yaw) are adjusted independently based on task performance, measured by an exponential moving average (EMA) of success.

Additionally, to prevent the \textit{Planner} from producing unstable manipulation behaviors, as well as redundant interactions, we introduce a smoothness regularization term inspired by L2C2~\cite{kobayashi2022l2c2}. For two consecutive states $o_{t}^{c}$ and $o_{t+1}^{c}$, a virtual state $\bar{o}_{t}^{c} = o_{t}^{c} + u(o_{t+1}^{c} - o_{t}^{c})$ is interpolated with $u \sim \mathcal{U}(0, 1)$. The smoothness loss is
\begin{equation}
\footnotesize
\mathcal{L}_{\text{smooth}} = 
\lambda_\pi \, D\!\left(\pi_\theta(s_t), \pi_\theta(\bar{s}_t)\right)
+ \lambda_V \sum_i D\!\left(V_{\phi_i}(s_t), V_{\phi_i}(\bar{s}_t)\right)
\vspace{-0.2cm}
\end{equation}
where $D$ denotes a distance metric, $\pi_\theta$ the actor, and $V_{\phi_i}$ the critics. This regularization enforces consistent outputs under minor state perturbations and reduces action jitter.


\section{Experiments}

\begin{table}[t]
\centering
\caption{Success rate and average completion time evaluated across 1000 simulations.}
\vspace{-0.3cm}
\setlength{\tabcolsep}{2.5pt} 
\label{tab:Success_rate}
\scriptsize
\begin{tabular}{@{}cccc@{}}
\toprule
\textbf{Method} & \textbf{Success Rate(\%)} & \textbf{Average Time(s)} \\ \midrule
\textbf{HeLoM w/o curriculum} & 49.3 & 21.4s \\ 
\textbf{HeLoM w/o Key reward} & 63.7 & 18.31s \\ 
\textbf{HeLoM w/o $\lambda_{\text{turn}}$ \& $\lambda_{\text{push}}$} & 76.5 & 16.9s \\ 
\textbf{HeLoM w/o smoothness loss} & 89.6 & 17.2s \\ 
\textbf{HeLoM} & \textbf{95.6} & \textbf{12.5s} \\ \bottomrule
\end{tabular}
\end{table}

We build our training environment in the Isaac Gym simulator and utilize one RTX 4090 GPU to train 4096 robot instances in parallel, which significantly accelerates the convergence of the RL process.
We then deploy the trained policy on a \textbf{HEBI Daisy} modular series-elastic hexapod robot, to validate performance in the real world.


\subsection{Simulation Results and Success Rate}

We randomly sample 1000 simulated interaction environments, under the training parameter range described in Section~\ref{sec:planner_training}. The physical properties of the objects are summarized in Table~\ref{tab:obj_size}. To quantitatively evaluate the effectiveness of \textbf{HeLoM}, we consider two metrics: \emph{success rate} and \emph{average completion time}. Task success is defined as pushing the object to the target within $25\,\text{s}$, with distance $d < 0.05\,\text{m}$ and angular error $\theta_z < 5^\circ$. Any trial exceeding this time limit is regarded as a failure. We compare the performance of our method with the following baselines:

\begin{itemize}
    \item \textbf{HeLoM w/o curriculum}, where we removed the command-based curriculum to evaluate the role of progressive task scaling in enabling learning over a large task space.

    \item \textbf{HeLoM w/o Key reward}, where we removed the rotation guidance term $r_{\text{obj,turn}}$ to evaluate its contribution to large-angle manipulation capability.

    \item \textbf{HeLoM w/o $\lambda_{\text{turn}}$ \& $\lambda_{\text{push}}$}, where we removed the phase-dependent weighting between rotation and translation to evaluate the effect of staged task decomposition on coordinating these objectives.

    \item \textbf{HeLoM w/o smoothness loss}, where we removed the smoothness regularization to evaluate its impact on efficiency and behavioral stability.
\end{itemize}

\begin{figure}[t] 
    \centering
    \includegraphics[width=\columnwidth]{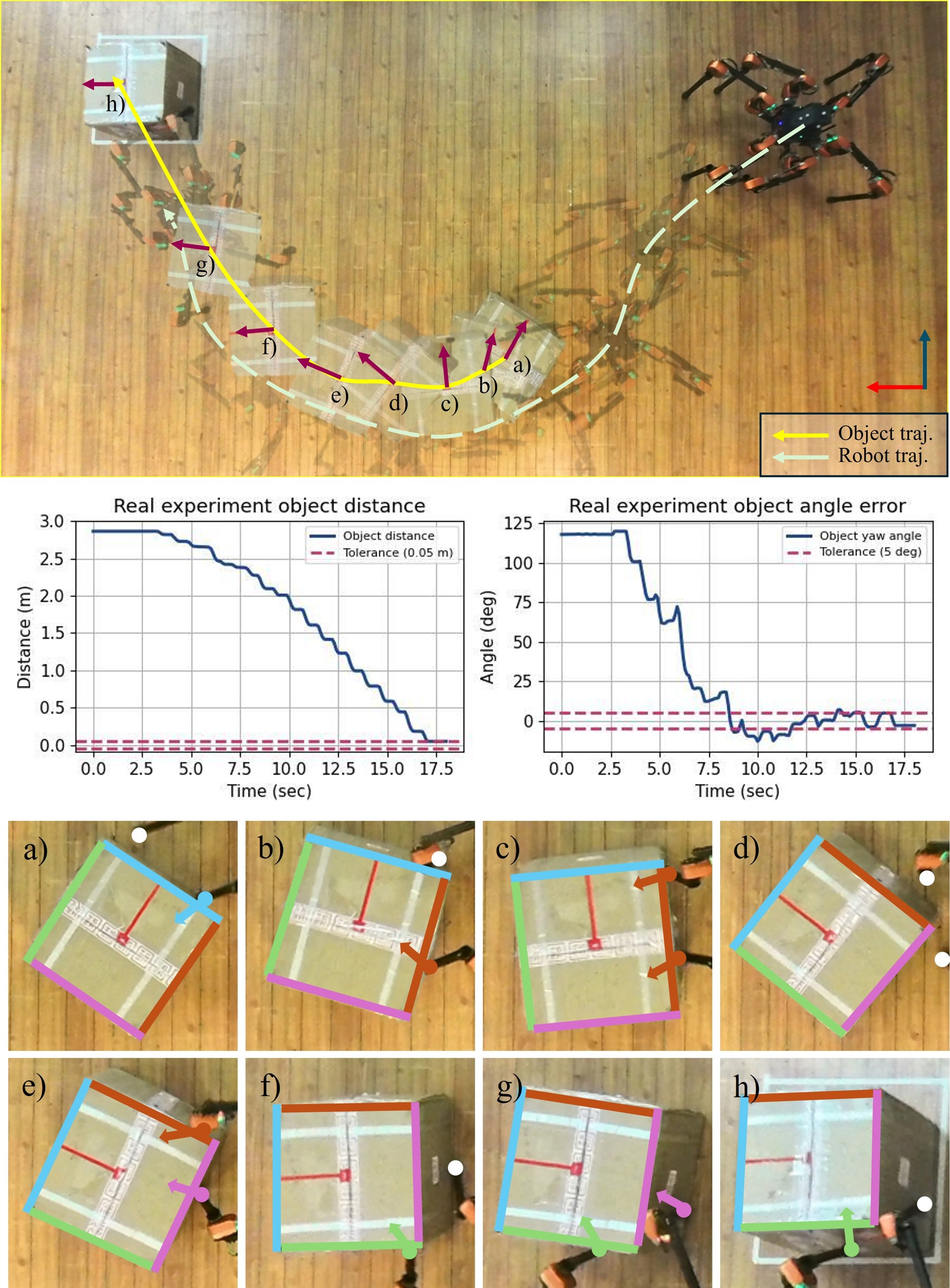}
    \vspace{-0.75cm}
    \caption{\textbf{Complete robot–object interaction process in a real-world trial}. Top: Robot and object trajectories with intermediate states (a–h).Middle: Object–goal distance and yaw error over time. Bottom: Snapshots of the manipulation process, white dots indicate no contact, while colored arrows denote both the contact surface and the direction of foot motion. }
    \vspace{-0.6cm}
    \label{fig:interaction}
\end{figure}

\begin{table}
\centering
\caption{Tracking comparison across command ranges.}
\vspace{-0.3cm}
\label{tab:tracking_comparison}
\renewcommand{\arraystretch}{1.25}
\resizebox{\columnwidth}{!}{
\begin{tabular}{lcccc}
\toprule
\textbf{Metric} & \textbf{Ranges} & \textbf{HeLoM $\mu_{|e|} \; [\min(e), \max(e)]$} & \textbf{PD only} & \textbf{AcNet only} \\
\midrule

\multirow{2}{*}{Linear velocity}
& [-0.2, 0.2] m/s & \textbf{0.035 \,[{-0.084},\,0.106]} & 0.044 \,[{-0.144},\,0.128] & 0.040 \,[{-0.171},\,0.174] \\
& [-0.4, 0.4] m/s & \textbf{0.044 \,[{-0.138},\,0.157]} & 0.067 \,[{-0.176},\,0.185] & - \\

\midrule

\multirow{2}{*}{Angular velocity}
& [-0.50, 0.50] rad/s & \textbf{0.105 \,[{-0.220},\,0.256]} & 0.146 \,[{-0.330},\,0.311] & 0.204 \,[{-0.383},\,0.357] \\
& [-0.75, 0.75] rad/s & \textbf{0.128 \,[{-0.282},\,0.302]} & 0.164 \,[{-0.357},\,0.429] & - \\

\midrule

\multirow{3}{*}{Foot tracking}
& $x$ [0.475, 0.725] m & \textbf{0.022 \,[{-0.081},\,0.016]} & 0.039 \,[{-0.018},\,0.103] & 0.048 \,[{-0.101},\,0.090] \\
& $y$ [-0.300, 0.300] m & \textbf{0.010 \,[{-0.024},\,0.025]} & 0.020 \,[{-0.021},\,0.024] & 0.043 \,[{-0.088},\,0.190] \\
& $z$ [-0.075, 0.075] m & \textbf{0.008 \,[{-0.011},\,0.012]} & 0.010 \,[{-0.018},\,0.026] & 0.070 \,[{-0.122},\,0.051] \\

\bottomrule
\end{tabular}
}
\vspace{-0.6cm}
\end{table}

\begin{figure*}[t]
    \centering
    \includegraphics[width=1.0\textwidth]{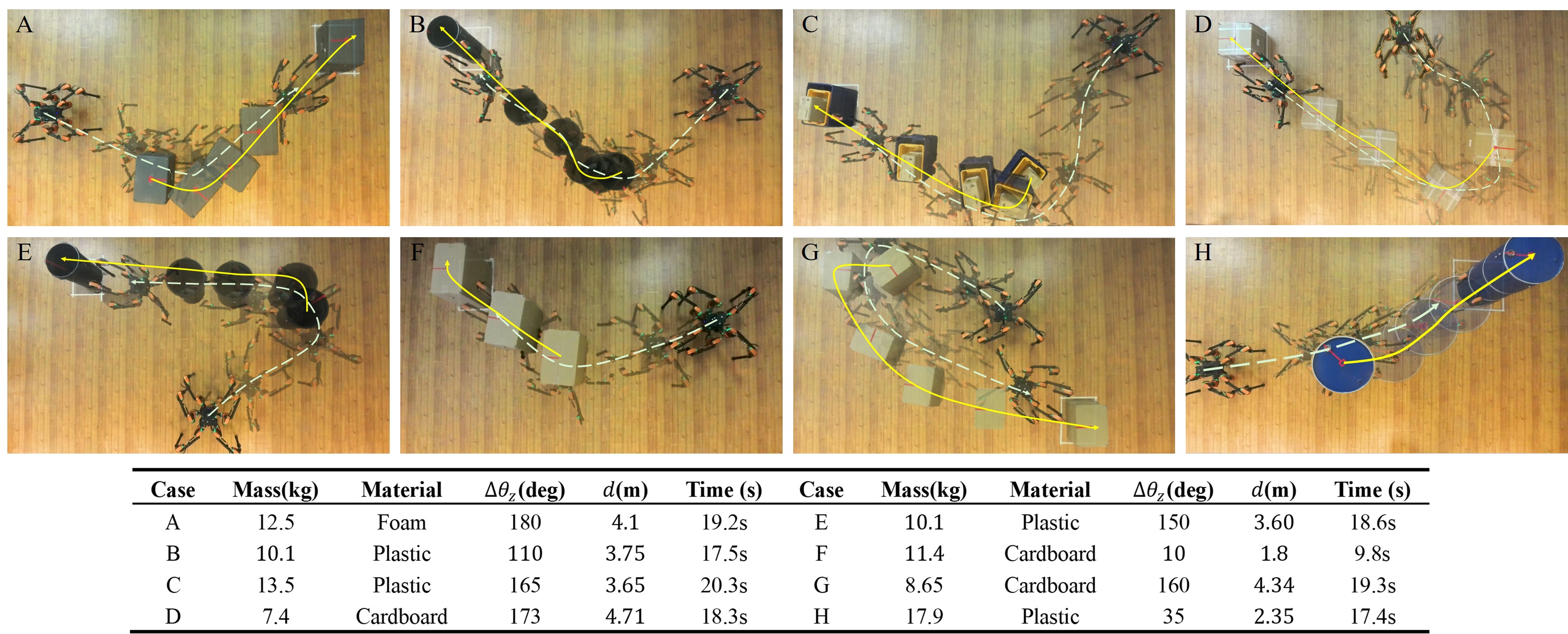}
    \vspace{-0.8cm}
    \caption{\textbf{Eight representative scenarios from real-world experiments}. The figure illustrates how the hexapod robot uses its legs to push objects with different initial poses, sizes, weights, materials, shapes, and centers of mass to the desired pose. The table below summarizes the object specifications and experimental parameters.}
    \label{fig:pushing}
\vspace{-0.6cm}
\end{figure*}

As shown in Table~\ref{tab:Success_rate}, \textbf{HeLoM} achieves the highest success rate of 95.6\% in these scenarios, while also attaining the shortest average completion time. Failures are mainly observed in lightweight, low-friction cases, where overshooting may occur due to excessive momentum.
In contrast, \textbf{HeLoM w/o Curriculum} yields the lowest success rate of only 20\%. Without progressive task scaling, the policy must directly learn over a large and diverse task space. This makes exploration highly challenging and often leads the policy to local optima. As a result, the robot only acquires simple contact and pushing behaviors, and is limited to straightforward, near-linear tasks. For \textbf{HeLoM w/o $r_{\text{obj,turn}}$}, removing the rotation guidance reward significantly degrades performance in tasks requiring large orientation changes. When the orientation error is large, relying solely on the orientation reward provides insufficient guidance for learning effective rotational manipulation behaviors. For \textbf{HeLoM w/o $\lambda_{\text{turn}}$ \& $\lambda_{\text{push}}$}, removing the staged weighting between rotation and translation introduces conflicts between the two objectives. For instance, rotating the object to reduce the orientation error may temporarily increase its distance to the target, leading to inconsistent learning signals. This often causes the robot to deviate from the desired trajectory or fail to reach the target pose. Finally, \textbf{HeLoM w/o smoothness loss} tends to generate redundant and inefficient manipulation behaviors, significantly increasing the average completion time. The robot often performs unnecessary adjustments or erratic motions, which destabilize object movement and prolong the alignment process.


\subsection{Hardware Validation}

\subsubsection{Loco-manipulation performance}
In our real-world experiments, we first evaluated the locomotion performance and manipulation accuracy of our \textit{Controller}. To further validate the effectiveness of the controller design, we compare it with two baseline methods. Table~\ref{tab:tracking_comparison} illustrates the tracking performance in both commanded velocity and foot position.

The results show that the proposed controller, which combines ACNet for locomotion and a PD controller for manipulation, achieves higher accuracy across all metrics compared to a pure PD controller. In contrast, a pure ACNet-based controller performs significantly poorer in manipulation accuracy and is unusable for high-velocity motions due to aggressive jittering in the front legs. This is consistent with prior work~\cite{xie2021dynamics}, which finds that sim-to-real transfer is highly sensitive to controller structure, and that controllers unsuited to the specific task can produce stiff or unstable behaviour. Therefore, PD control is retained for manipulation, where explicit position regulation is required, while ACNet is applied only to the locomotion legs where accurate actuator modelling is more beneficial.

\subsubsection{Legs Interaction with Unknown Objects}

For our pushing tasks, we employed an external optical motion capture system to acquire the accurate poses of both the robot and the object in the world frame. The current and target pose of the object is transformed into the robot's local body frame and used as input to the policy.

Fig.~\ref{fig:interaction} illustrates the motion states and interaction process of the robot when pushing an unknown object. Throughout the process, the robot relies only on the perceived relative poses of the object and the target (in the robot frame), together with proprioceptive feedback, to guide its leg-driven interactions and accomplish the pushing task. In particular, as shown in Fig.~\ref{fig:interaction}(a–h), the robot further switches contact surfaces at key moments and adjust both the contact locations and motion directions, enabling continuous correction of the object's orientation during pushing. This sequence of actions demonstrates the robot's ability to adaptively select interaction strategies when pushing unknown objects, thereby achieving stable and efficient manipulation. Moreover, we observe that, benefiting from coordinated multi-limb interaction, the robot can accomplish efficient object relocation with only minor adjustments to its own posture, which can be clearly observed from the smooth robot trajectory. Meanwhile, both the object–goal distance and yaw error decrease steadily over time and converge within tight tolerances ($d < 0.05\,\text{m}$ and $\theta_z < 5^\circ$).

\subsubsection{Whole-body Pushing Performance and Robustness}

\begin{figure}[t] 
    \centering
    \includegraphics[width=\columnwidth]{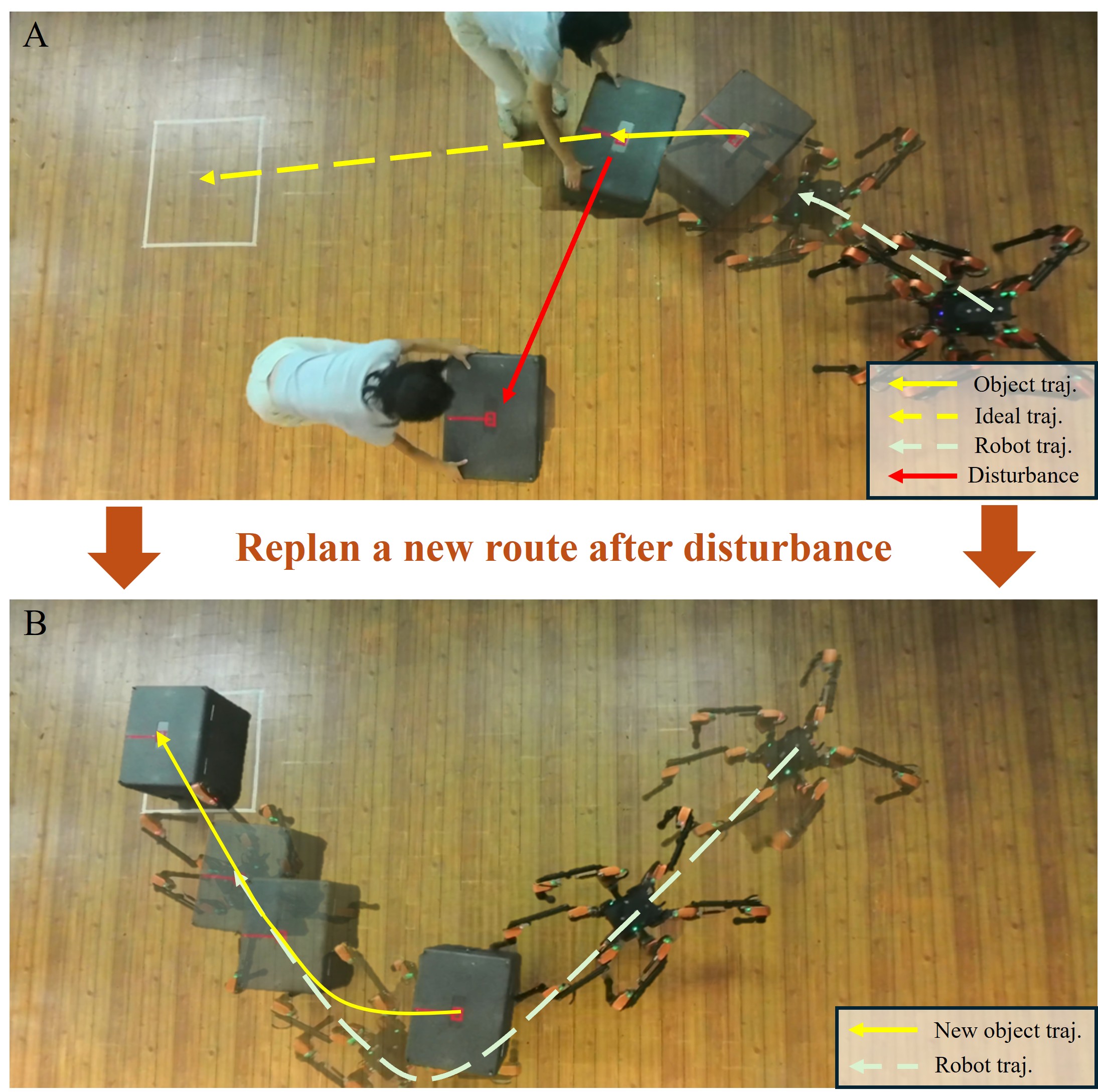}
    \vspace{-0.8cm}
    \caption{\textbf{Human intervention introduced during the robot pushing task}. (A) A disturbance is introduced by manually shifting the object away from its intended trajectory, forcing the robot to deviate from its current path. (B) The robot adapts by replanning a new route, coordinating its forelegs and body to realign the object and guide it back toward the target position.}
    \vspace{-0.6cm}
    \label{fig:disturbance_test}
\end{figure}

To demonstrate the generalization capability of \textbf{HeLoM}, we select eight representative scenarios, as shown in Fig.~\ref{fig:pushing}. The robot is initialized at arbitrary positions and orientations to push boxes and cylinders from different directions, including forward, backward, and lateral pushing. Our experimental results show that \textbf{HeLoM} can effectively handle objects with varying heights, weights, and materials. Even in challenging cases such as Fig.~4(a), where the center of mass is significantly shifted toward a corner, resulting in uneven mass and friction distribution, the robot can still complete the task with stable performance. Case (d) further involves an uneven surface not encountered during training, which requires precise foot placement to maintain effective contact. If the positioning is not well executed, the foot may become stuck or lose effective contact, which makes the pushing process significantly more challenging. Furthermore, we also evaluate objects beyond the training distribution in terms of weight and size, as shown in Fig.~\ref{fig:pushing}(h). In addition, we conduct a load-adaptive whole-body pushing test to highlight the capability of \textbf{HeLoM} in coordinating whole-body behaviors. As the load increases, the robot lowers its body and adjusts its posture to maintain stable and continuous pushing. More details can be found in the accompanying video.

To further evaluate the robustness of the proposed policy in uncertain environments, we introduced external disturbances during the robot's pushing process to reflect real-world variability. As shown in Fig.~\ref{fig:disturbance_test}, we manually perturbed the object's position and orientation while the robot was executing the task. The robot adapted its strategy, replanned its actions, and still completed the task successfully. These results show the strong adaptability and resilience of our proposed framework in handling unexpected disturbances.


\section{Conclusion}

In this paper, we introduce \textbf{HeLoM}, a whole-body manipulation framework for hexapod robots.
By fully leveraging all legs, our framework enables stable and robust interactions with a wide range of objects.
\textbf{HeLoM} enables the robot to achieve coordinated whole-body behaviors by adjusting its body posture together with the positions and motion directions of its feet, allowing it to generate effective pushing actions that drive the object toward the desired pose.
However, our current implementation relies on an external motion capture system for precise state feedback, which limits its use to controlled environments. 
Integrating onboard sensing modules such as cameras and LiDAR may allow us to overcome these constraints and enable deployment to real-world scenarios.
That being said, we believe that the results of this study not only validate the effectiveness and potential of multi-legged robots as stable and versatile manipulation platforms but also demonstrate that decomposing whole-body manipulation tasks into planning and control provides a clear task allocation and an efficient execution mechanism for solving complex manipulation problems.


\bibliographystyle{IEEEtran}
\bibliography{IEEEexample}

\end{document}